\documentclass{article}

\PassOptionsToPackage{numbers}{natbib}

\usepackage[preprint]{nips_2018}

\usepackage[utf8]{inputenc} 
\usepackage[T1]{fontenc}    
\usepackage{hyperref}       
\usepackage{url}            
\usepackage{booktabs}       
\usepackage{amsfonts}       
\usepackage{nicefrac}       
\usepackage{microtype}      
\usepackage{amsmath, amsthm, amssymb, graphicx, subfigure}
\newcommand{\ra}[1]{\renewcommand{\arraystretch}{#1}}
\newcommand{\argmax}{\operatornamewithlimits{argmax}}

\newtheorem{thm}{Theorem}

\newtheorem{col}{Corollary}

\def\kl{\mathop{\rm KL}\nolimits}
\def\hel{\mathop{\rm H}\nolimits}
\def\tr{\mathop{\rm tr}\nolimits}
\newcommand{\cqfd}{\tag*{$\blacksquare$}}

\title{Adversarial confidence and smoothness regularizations for scalable unsupervised discriminative learning}

\author{
  Yi-Qing Wang\\
  School of Computer Science and Engineering\\
  Nanyang Technological University\\
  Singapore\\
  \texttt{yqwang9@gmail.com} \\
}

\begin{document}

\maketitle
\begin{abstract}
In this paper, we consider a generic probabilistic discriminative learner from the functional viewpoint and argue that, to make it learn well, it is necessary to constrain its hypothesis space to a set of non-trivial piecewise constant functions. To achieve this goal, we present a scalable unsupervised regularization framework. On the theoretical front, we prove that this framework is conducive to a factually confident and smooth discriminative model and connect it to an adversarial \textit{Taboo} game, spectral clustering and virtual adversarial training. Experimentally, we take deep neural networks as our learners and demonstrate that, when trained under our framework in the unsupervised setting, they not only achieve state-of-the-art clustering results but also generalize well on both synthetic and real data.
\end{abstract}

\section{Introduction} \label{sec:intro}

Classification is a longstanding field of study in statistical learning. In the supervised setting where both instances and their labels are available, a solid theoretical foundation \cite{devroye1996probabilistic} has been established to guide the design of an effective learning algorithm. A key piece of this theory concerns the hypothesis space. To ensure successful supervised learning with a limited amount of labeled data, it has to be neither too large nor too small. In reality, this theoretical dilemma has been played out to the full. Just a few decades earlier, the hypothesis spaces built for the machines were too small to capture meaningful concepts. They underfitted. Deep learning has fundamentally changed the landscape, in vision related tasks in particular \cite{lecun1998gradient} \cite{krizhevsky2009learning}, because of their ability to learn sophisticated features automatically. Equipped with these highly expressive hypothesis spaces, researchers now face the opposite issue, i.e. how to take full advantage of them without causing overfitting, since labeled data typically requires human annotation and is thus in limited supply. Its solution, from the statistical learning theory's viewpoint, is clear: regularization. 


In this paper, we take the functional view of a generic probabilistic discriminative learner and propose a scalable unsupervised framework for its regularization. The core idea is to constrain its hypothesis space to a set of non-trivial piecewise constant functions. It can be motivated as follows. In most classification tasks, for a crushing majority of instances, one is often unequivocal regarding the category to which they should be assigned. This \textit{factual confidence}, translated mathematically, means that with overwhelming probability, the true predictive distribution conditional on an instance is a unit mass. As a function, it is thus approximately discrete-valued. Furthermore, frequently, our certainty regarding the label of an instance is such that even when altered to some degree, the instance will still be recognized as a member from the same category. Interpreted again in terms of the true conditional distribution, this property suggests its \textit{smoothness}. As a result, if a probabilistic discriminative model is to match its learning target, namely the true conditional distribution, it is naturally necessary for it to be \textit{factually} confident and smooth, hence \textit{non-trivially} piecewise constant. 

To encourage a model to be non-trivially confident in its predictions, we draw inspiration from a familiar \textit{Taboo} game. We analyse the game as a communication process in an adversarial environment and derive from it a scalable unsupervised regularization surrogate, whose minimization leads to a factually confident discriminator. Hence referred to as \textit{confidence regularization}, it can be shown as a scalable approach to spectral clustering \cite{von2007tutorial}. 

To ensure a model's smoothness, we introduce an unsupervised \textit{smoothness regularization}. Its main inspiration is \textit{virtual adversarial training} \cite{miyato2017virtual}. In particular, the functional view of a discriminator allows us to derive the loss rigorously from a uniform smoothness property. It also leads to a criterion for measuring a discriminator's local predictive stability. Moreover, our study suggests that confidence and smoothness are not isolated properties, in that when a smooth discriminator is confident of its predictions, under certain conditions, it becomes immune to attacks \cite{goodfellow2014explaining} and able to generalize.



In the rest of the paper, we begin by formulating the \textit{Taboo} game, which leads to the first regularization loss that induces a non-trivially discrete-valued discriminator. Next we define the smoothness for a discriminative model and derive the second regularization loss in Section \ref{sec:vat}. Then we validate our approach with some experiments on unsupervised discriminative learning in Section \ref{sec:experiments}. Finally, we conclude by relating our framework to prior works and discussing future research directions. 

\paragraph{Notations.} For some positive integer $d \in \mathbb{Z}_+$, $\mathbb{R}^d$ denotes the $d$-dimensional Euclidean space and $\|\cdot\|_p$ its $p$-norm. Assume w.l.o.g. the instance and label space $(\mathcal{X}, \mathcal{Y}) \subseteq \mathbb{R}^h\times \mathbb{Z}_+$. Consider a random pair $(X, Y) \in \mathcal{X} \times \mathcal{Y}$. Let $\mathbb{Q}^*$ denote its true distribution and $\mathbb{Q}(\cdot|X; \theta)$ a model for $\mathbb{Q}^*(\cdot|X)$ with parameter $\theta\in \mathbb{R}^q$. Denote by $c^*: \mathcal{X} \mapsto \mathcal{Y}$ the Bayes classifier i.e. $c^*(x) := \argmax_{y\in \mathcal{Y}} \mathbb{Q}^*(y|x)$ with ties broken arbitrarily. For any label $y \in \mathcal{Y}$, we denote the collection of all $y$-labeled instances by $\mathcal{X}_y := \{x \in \mathcal{X}\;| \; c^*(x) = y\}$. The indicator of any set $\mathcal{A}$ is denoted by $1_{\mathcal{A}}(z)$. It equals $1$ if $z\in \mathcal{A}$ and $0$ otherwise. Finally, $f(\cdot||\cdot)$ denotes an f-divergence that measures the discrepancy of two laws. 

\section{From \textit{Taboo} to a factually confident discriminator} \label{sec:taboo}

Our \textit{Taboo} game involves three players, Ann, Bob and Cal. Ann plays an adversarial role against Bob and Cal. Assume a fixed number of categories known to all of them. The game goes as follows. First, Ann gives Bob a diverse collection of unlabeled instances. Next she selects a category, reveals it to Bob, and asks him to describe it to Cal using a single instance from his collection. Bob and Cal win the game if Cal, upon receiving the instance, is able to correctly identify the category. We now formulate Bob and Cal's interaction as a communication process and show that if they have to win regardless of how Ann picks the initial category, they must be confident in their moves. 

\paragraph{\textit{Taboo} as label transmission.} Consider a discriminative model $\mathbb{Q}(\cdot|x), \; x \in \mathcal{X}$ and a finite instance set $\mathcal{S} \subseteq \mathcal{X}$. We model the communication between Bob and Cal with the \textit{label transition matrix} 
\begin{align}
	\forall (y, y') \in \mathcal{Y}\times\mathcal{Y}, \quad
	\mathbb{T}(y'|y; \mathbb{Q}, \mathcal{S}) := \sum_{x \in \mathcal{S}} \mathbb{P}(x|y; \mathbb{Q}, \mathcal{S}) \mathbb{Q}(y'|x) \label{labeltm}
\end{align}
where
\begin{align}
\mathbb{P}(x|y; \mathbb{Q},\mathcal{S}) := 1_{\mathcal{S}}(x)\frac{\mathbb{Q}(y|x)}{\sum_{x' \in \mathcal{S}} \mathbb{Q}(y|x')}. \label{l2i}
\end{align}
For any $y \in \mathcal{Y}$ such that $\sum_{x' \in \mathcal{S}} \mathbb{Q}(y|x') = 0$, the probability $\mathbb{P}(x|y; \mathbb{Q},\mathcal{S})$ can be defined arbitrarily. 


Specifically, Eq. \eqref{labeltm} and \eqref{l2i} describe how Bob, endowed with the model $\mathbb{Q}(\cdot|x)$ and the set $\mathcal{S}$, selects a single unlabeled instance to convey the label information to Cal, who then decodes using the same $\mathbb{Q}(\cdot|x)$. Since the matrix $\mathbb{T}$ quantifies the likelihood for Cal to correctly infer the label intended by Bob, an ideal communication requires it to be an identity. Intuitively, a factually confident discriminator $\mathbb{Q}(\cdot|x)$ could meet this requirement. Because then Bob would \textit{only} select instances that are representative of the category chosen by Ann, whereas Cal would not confuse a received instance with one from another, different category. The next theorem shows that it is indeed the case.


\begin{thm}
Consider an arbitrary finite set $\mathcal{S} \subseteq \mathcal{X}$. The transition matrix $\mathbb{T}(\cdot|\cdot; \mathbb{Q}, \mathcal{S})$ is diagonal if and only if for all $x \in \mathcal{S}$, $\mathbb{Q}(\cdot|x)$ is valued in $\{0, 1\}$ and $\forall y \in \mathcal{Y}, \; \{x \in \mathcal{S}\;| \; \mathbb{Q}(y|x) = 1\}\neq \emptyset$.
\label{thm1}
\end{thm}
Hence, perfect label transmission through a single instance implies a confident model $\mathbb{Q}(\cdot|x)$ which additionally has to partition $\mathcal{S}$ into $|\mathcal{Y}|$ components, i.e. this confidence is fact based and not blind. This result suggests a necessary condition for a model $\mathbb{Q}(\cdot|x)$ to match its learning target i.e. $\mathbb{Q}^*(\cdot|x)$.
\begin{col}
Assume a classification setup with zero Bayes error i.e. $\mathbb{Q}^*(c^*(X) \neq Y) = 0$. $\mathbb{Q}(\cdot|x) = \mathbb{Q}^*(\cdot|x), \forall x\in \mathcal{X}$ holds only if the transition matrix $\mathbb{T}(\cdot|\cdot; \mathbb{Q}, \mathcal{S})$ is diagonal for any set $\mathcal{S} \subseteq \mathcal{X}$ such that $\forall y \in \mathcal{Y}, \;\mathcal{S} \cap \mathcal{X}_y  \neq \emptyset$ i.e. $\mathcal{S}$ contains at least one instance from every category in $\mathcal{Y}$.\label{col1}
\end{col}
All the proofs are in the Appendix. Zero Bayes error assumption, also called the \textit{noiseless} condition, is common for analyzing learning algorithms in classification \cite{devroye1996probabilistic}. In the following, a set that has at least one representative for each \textit{true} category will be referred to as \textit{label complete}. 

\paragraph{Confidence regularization.} Hence, a good discriminator should allow for successful transmission of \textit{any} label with \textit{any} label complete set. This observation leads to the confidence regularization. Specifically, consider an unlabeled set $\mathcal{U}$. Define the collection of its label complete subsets as 
\begin{align*}
	\Lambda(\mathcal{U}) := \{\mathcal{S} \subseteq \mathcal{U} | \; \forall y \in \mathcal{Y}, \; \mathcal{S} \cap \mathcal{X}_y \neq \emptyset\}. 
\end{align*}
In view of Theorem \ref{thm1} and Corollary \ref{col1}, denoting by $1_y$ the discrete probability over $\mathcal{Y}$ with a unit mass on $y$, we define the following loss for a parametrized discriminative model $\mathbb{Q}(\cdot|x; \theta)$
\begin{align}
	L_c(\theta;\mathcal{U}):=\max_{(\mathcal{S}, y)\in \Lambda(\mathcal{U}) \times \mathcal{Y}}f (1_y ||\mathbb{T}(\cdot|y; \theta, \mathcal{S})).  \label{uform0} 
\end{align}
Its interpretation in the \textit{Taboo} game is simple: knowing Bob and Cal's discriminative model, Ann picks a label complete set and a category so as to make it as hard as possible for them to conduct a successful communication. Eq. \eqref{uform0} quantifies the resulting worst-case label transmission failure rate. 

Since the subsets $\mathcal{X}_y, y\in \mathcal{Y}$ are unknown, so is $\Lambda(\mathcal{U})$. As a result, Eq. \eqref{uform0} is impractical. But if we can somehow sample from $\Lambda(\mathcal{U})$ according to some law $\mathbb{B}$, Eq. \eqref{uform0} can be relaxed to
\begin{align}
	L'_c(\theta;\mathbb{B}):=\mathbb{E}_{\mathcal{S}\sim \mathbb{B}(\Lambda(\mathcal{U}))}\left[\max_{y\in \mathcal{Y}}f (1_y ||\mathbb{T}(\cdot|y; \theta, \mathcal{S}))\right].  \label{uform1} 
\end{align}
Compared to Eq. \eqref{uform0}, this loss assumes a friendlier Ann in that she now selects Bob's label complete set at random according to $\mathbb{B}(\Lambda(\mathcal{U}))$, after which she still picks a category adversarially. 

Sampling label complete sets is indeed possible. To see it, consider a partition of $\Lambda$: $\cup_{b \geq |\mathcal{Y}|} \Lambda_b$ with $\Lambda_b := \{\mathcal{S} \in \Lambda\;|\;  |\mathcal{S}| = b\}$. The next theorem shows that for any sufficiently large $b$, we can reliably sample from $\Lambda_b$, hence $\Lambda$, by simply putting together $b$ random training instances.\begin{thm}
	Assume $\min_{y\in \mathcal{Y}}\mathbb{Q}^*(Y=y) > 0$. Let $(\mathcal{S}_j)_{j=1,\ldots,T}$ be unlabeled sets of the same size $b$ consisting of $\mathbb{Q}^*$-i.i.d. instances. For any $\epsilon \in (0,1)$, if $b > \ln \left(T |\mathcal{Y}|\epsilon^{-1}\right)/\min_{y\in \mathcal{Y}}\mathbb{Q}^*(Y=y)$,
then $\mathbb{Q}^*\left(\cap_{t=1}^T\{\mathcal{S}_t \in \Lambda_b\}\right) > 1- \epsilon$, i.e. with probability at least $1-\epsilon$, all $T$ sets are label complete. \label{thm2} 
\end{thm}



See the Appendix for its proof. As an illustration, consider a balanced dataset $\mathcal{U}$. To approximately sweep it $r$ times and ensure with probability at least $1-\epsilon$ that all the random batches sampled in the process are label complete, the batch size $b$ needs to satisfy $b > |\mathcal{Y}|\ln \left(r|\mathcal{U}||\mathcal{Y}|\epsilon^{-1}b^{-1}\right)$. For $|\mathcal{Y}|=10, |\mathcal{U}|=6\times10^4, \epsilon=10^{-4}$ and $r=1000$, it implies $b\geq240$.
	
As a result, we sample label complete sets in this way and will refer to the \textit{confidence regularization} loss Eq. \eqref{uform1} as $L_c'(\theta; \mathcal{U}, b)$ with some training batch size $b$ set according to Theorem \ref{thm2}. As a side note, confidence regularization can be seen as a \textit{scalable} approach to spectral clustering \cite{von2007tutorial} and is connected to association learning \cite{haeusser2017learning}. We defer this discussion to Section \ref{sec:relation}.

\section{Model smoothness and immunity to adversarial attacks} \label{sec:vat}

Even when confidence regularized, a complex model can still be erratic. To see it, assume $\mathcal{Y}=\{0,1\}$. Then an arbitrary definition $\mathbb{Q}(Y=1|\cdot): x\in\mathcal{X} \mapsto \{0,1\}$ such that $\mathbb{E}_{X\sim \mathbb{Q}^*}[\mathbb{Q}(Y=1|X)]=1/2$ (i.e. high entropy regime) is likely to result in a confident yet jittery model. We want to avoid that.


Enter the smoothness requirement. A discriminative model is smooth if its output, a distribution over $\mathcal{Y}$, varies continuously as one moves around in the instance space $\mathcal{X} \subseteq \mathbb{R}^h$. Formally, in the noiseless setting, for some f-divergence $f(\cdot||\cdot)$, a model $\theta$ is said to be smooth w.r.t. $\mathbb{Q}^*$ if it satisfies
\begin{align}\text{for $\mathbb{Q}^*$-almost all instances $x$}, \quad
\lim_{\rho \to 0}\sup_{\|r\|_2 \leq \rho}f(\mathbb{Q}(\cdot|x; \theta) || \mathbb{Q}(\cdot|x+r; \theta)) =0\label{continuity} 
\end{align}
where $r \in \mathbb{R}^h$ is a perturbation vector. Note that if the instance space $\mathcal{X}$ is a manifold in $\mathbb{R}^h$, at some $x$, it may be more appropriate to use a subset of the $\ell_2$-ball $\{r\;| \; \|r\|_2\leq \rho\}$ in the definition. However, since an instance space is typically hard to describe analytically and that it may additionally be task dependent, the $\ell_2$-ball is a convenient choice out of the worst-case consideration. 


\paragraph{Smoothness regularization.} If the true conditional distribution $\mathbb{Q}^*(\cdot|x), x \in \mathcal{X}$, as has been argued in the introduction, is piecewise constant, not only does it satisfy Eq. \eqref{continuity}, which is a local uniform smoothness property, it is also globally uniformly smooth w.r.t. $\mathbb{Q}^*$. This observation leads us to define $\mathbb{Q}^*$'s \textit{attack-free margin} as
\begin{align}
	\rho^* :=\sup\left\{\rho \;\Big| \; \text{for $\mathbb{Q}^*$-almost all instances $x$}, \; \sup_{\|r\|_2 \leq \rho}f(\mathbb{Q}^*(\cdot|x) || \mathbb{Q}^*(\cdot|x+r)) =0\right\}. \label{radius} 
\end{align}
It measures the maximum extent to which \textit{any} $\mathbb{Q}^*$-a.s. instance can be perturbed without causing \textit{true} confidence decline. In this light, a new necessary condition arises for a discriminative model $\theta$ to match its learning target. Specifically, under $\mathbb{Q}^*$, given an unlabeled set $\mathcal{U}$ and an f-divergence, we define the next \textit{smoothness regularization} loss to assess a model $\theta$'s global uniform smoothness
\begin{align}
	L_s(\theta;\mathcal{U}, \rho^*) := \max_{x\in \mathcal{U}} \sup_{\|r\|_2\leq\rho^*}f(\mathbb{Q}(\cdot|x; \theta) || \mathbb{Q}(\cdot|x+r; \theta)). \label{cont0} 
\end{align}
But it is impractical since 1) the attack-free margin $\rho^*$, a property of $\mathbb{Q}^*$, is unknown in general and 2) the maximum over the instances makes it hard for a learner to scale. Hence, we relax Eq. \eqref{cont0} to 
\begin{align}
	L'_s(\theta;\mathcal{U}, \rho) := |\mathcal{U}|^{-1}\sum_{x\in \mathcal{U}} \sup_{\|r\|_2\leq\rho}f(\mathbb{Q}(\cdot|x; \theta) || \mathbb{Q}(\cdot|x+r; \theta)) \label{cont1}
\end{align}
where the maximum over the instances is replaced by a sample average and the attack-free margin by a tunable parameter $\rho$ set by e.g. cross-validation. The remaining issue is to find the supremum in Eq. \eqref{cont1} or at least a good lower bound to it for an arbitrary instance $x$ and a positive $\rho$. 

To this end, denote $\phi_f(r; x, \theta):= f\left(\mathbb{Q}(\cdot|x; \theta) || \mathbb{Q}(\cdot|x+ r; \theta)\right)$. For some fixed unit vector $e \in \mathbb{R}^h$, consider $\psi_f(\nu; e, x, \theta) := \phi_f(\nu e; x, \theta)$ with $\nu \in \mathbb{R}$. It has some interesting properties. 
\begin{thm}
	Assume $\psi_f(\cdot; e, x, \theta)$ twice differentiable around $\nu=0$. Then both its value and first derivative vanish at $\nu=0$ i.e. $\psi_f(0; e, x, \theta)  = \psi_f'(0; e, x, \theta) = 0$. In addition, when the f-divergence is the Kullback-Leibler divergence $\kl(\cdot||\cdot)$ or the squared Hellinger distance $\hel^2(\cdot||\cdot)$, we find 
\begin{align*}
	\psi''_{f}(0; e, x, \theta) = c_f e^t \left(\sum_{y \in \mathcal{Y}} \mathbb{Q}(y|x;\theta) \nabla_x \log  \mathbb{Q}(y|x; \theta) \nabla^t_x \log  \mathbb{Q}(y|x;\theta)\right) e := c_f e^t I_F(x, \theta) e
\end{align*}
with $I_F(x,\theta)$ the model $\theta$'s Fisher information at $x$ and $c_f$ a constant: $c_{\kl}=1$ and $c_{\hel^2}=1/4$. \label{lem1}
\end{thm}
Its proof (omitted due to page limit) follows directly from the definition of an f-divergence. This result is natural given the Fisher information's instrumental role in estimator design e.g. the Cramer-Rao bound and information geometry \cite{amari1987differential}. From this perspective, Condition \eqref{continuity} may be interpreted as requiring that locally, the label $Y$ reveals as little information as possible about the instance $X$.  

To anticipate the following development, note that by construction, the Fisher information $I_F(x,\theta)$ has its rank upper bounded by $\min(|\mathcal{Y}|, h)$. In particular, at any instance $x$, let $A(x, \theta)$ be an $h\times |\mathcal{Y}|$ matrix whose $y$-th column corresponds to $\sqrt{\mathbb{Q}(y|x;\theta)} \nabla_x \log  \mathbb{Q}(y|x; \theta)$. Then $I_F(x,\theta) = A(x,\theta)A^t(x,\theta)$. 
	



Theorem \ref{lem1} implies that $\phi_f(\cdot; x,\theta)$ is approximately convex in an infinitesimal neighborhood around $r=0$. It further indicates that, at any $x$, to get a good lower bound to $\sup_{\|r\|_2\leq \rho}\phi_f(r;x,\theta)$ for a small but not infinitesimal $\rho$, we pay special attention to the subspace spanned by $I_F(x,\theta)$'s \textit{leading} eigenvectors. It is thus natural to sample $\phi_f(\cdot;x,\theta)$ along the dimensions emphasized by the Gaussian vector $\mathcal{N}(0, I^k_F(x,\theta))$ for some positive integer $k$. A larger $k$ puts more focus on eigendimensions corresponding to $I_F(x,\theta)$'s larger eigenvalues. This reasoning leads to the \textit{random} lower bound 
\begin{align}
	\max_{\nu\in A, 1\leq i\leq m}\phi_f(\nu\rho e_i(x, \theta)/\|e_i(x,\theta)\|_2; x, \theta), \; e_i(x,\theta) \sim \mathcal{N}(0, I_F^k(x,\theta)), \; i=1,\ldots, m
\end{align}
where $A$ denotes a finite search set such that $\{-1,1\} \subseteq A\subset[-1,1]$ and $e_i(x,\theta)$ $m$ i.i.d. samples. Due to the Fisher information $I_F(x,\theta)$'s particular structure, it costs $O(k|\mathcal{Y}|h)$, rather than $O(kh^2)$, to generate a sample $e_i(x,\theta)$, which can be obtained as $A(x,\theta) n_i$ for $k=1$ and $A(x,\theta)A^t(x, \theta)n'_i$ for $k=2$ etc. with $n_i$ and $n'_i$ a standard Gaussian vector sample. Hence, this random lower bound can be computed efficiently, especially in high dimension i.e. $h \gg |\mathcal{Y}|$. In practice, we take $m=1$, $k=4$ and $A$ a uniform grid, containing e.g. $10$ evenly spaced points. Since these choices yielded good experimental results, no further attempt was made to optimize these parameters. 

\paragraph{Fisher criterion for local stability and immunity to attacks.} Write the Fisher information's largest eigenvalue as $\beta(x,\theta)$. Theorem \ref{lem1} suggests it be used for assessing a model $\theta$'s curvature, or stability at instance $x$. Again, due to $I_F(x,\theta)$'s low-rankedness, it can be computed efficiently especially when $h \gg |\mathcal{Y}|$ because $A^t(x,\theta)A(x,\theta)$ has the same spectra as $I_F(x,\theta)$. However, a more appealing alternative is the trace as the two are equivalent: $\tr(I_F(x, \theta))\min(|\mathcal{Y}|, h)^{-1} \leq  \beta(x, \theta) \leq \tr(I_F(x, \theta))$. Henceforth, we refer to the trace as the \textit{Fisher criterion for local stability}.




Furthermore, Theorem \ref{lem1} suggests that an attack \cite{goodfellow2014explaining} might succeed at instance $x$ only if the perturbation is not orthogonal to the space spanned by the score vectors $\nabla_x \log  \mathbb{Q}(y|x;\theta), y \in \mathcal{Y}$. In particular, if a smooth discriminator is discrete-valued as we seek to achieve by confidence regularization, an attack would be difficult because all the score vectors are zero. This analysis further suggests that when well trained, a factually confident and smooth model should be able to generalize, especially if the true distribution $\mathbb{Q}^*$ it attempts to match admits a large attack-free margin $\rho^*$ i.e. Eq. \eqref{radius}. 

\section{Experiments} \label{sec:experiments}

For some positive hyper-parameters $(\lambda, \rho)$, our study leads to the unsupervised regularization loss $R(\theta;\mathcal{U}, b, \rho, \lambda):=L'_c(\theta; \mathcal{U}, b)+ \lambda L'_s(\theta; \mathcal{U}, \rho)$. Though in principle, it applies to all probabilistic discriminative learner, we took neural nets in the experiments due to their large hypothesis spaces. They had ReLU units and batch normalization \cite{ioffe2015batch} and were optimized by ADAM \cite{kingma2014adam} with a constant learning rate $10^{-3}$. The training batch size $b$ was set with the sampling failure rate $\epsilon=10^{-4}$. For the f-divergence, we took the squared Hellinger distance in smoothness regularization for its symmetry and the Kullback-Leiber divergence in confidence regularization. \textit{Unsupervised clustering accuracy} was used for result evaluation. For a set of $m$ instances, it is defined as $m^{-1}\max_{\pi}\sum_{i=1}^m 1_{y_i}(\pi(l_i))$ where $l_i$ denotes a model's predicted label for instance $i$ and the maximum is over all permutations of $\mathcal{Y}$. It was computed using the Munkres algorithm \cite{munkres1957algorithms}. The neural nets were written in \textit{Pytorch} with a fixed random seed $0$. To test their generalization ability, a network was trained only on training data. For clustering, both training and test data was used. We report the best results. Rival algorithms with accuracy reported in both mean $\eta$ and standard deviation $\sigma$, we provide them as $\eta + \sigma$.

\subsection{Synthetic data}

\begin{figure}
  \centering
\subfigure[][epoch 1]{\includegraphics[height = 0.09\textheight, width = 0.325\textwidth]{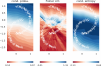}}
\subfigure[][epoch 200]{\includegraphics[height = 0.09\textheight, width = 0.325\textwidth]{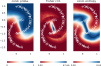}}
\subfigure[][epoch 1200]{\includegraphics[height = 0.09\textheight, width = 0.325\textwidth]{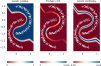}}
\caption{Unsupervised discriminative learning on a 2-c dataset. Each subfigure consists of (from left to right) model conditional probability, its Fisher criterion and conditional entropy. The model did end up being piecewise constant with a good model attack-free margin, which allows to generalize.} \label{fig0}
\end{figure}

To validate our approach, we tested it on a low noise 2-c dataset \cite{shaham2018spectralnet}. Both training and test data contain $300$ random points per class. We took a net of two fully connected hidden layers comprising respectively $200$ and $100$ neurons and trained it only on the training data with $\lambda=1000$ and $\rho=0.04$. Note that too large a $\rho$ requires the model to be smooth even for two points from different classes, provided that their Euclidean distance is smaller than $2\rho$. Too small a $\rho$ fails to enforce model smoothness among points of the same class, hence $\rho$'s alternative interpretation as \textit{neighborhood width}. Fig. \ref{fig0} shows that the trained net succeeded in clustering the test data. Moreover, it did end up being $\mathbb{Q}^*$-piecewise constant with a good model attack-free margin, which allows to generalize. 

\subsection{MNIST}

MNIST has roughly balanced training and test set containing respectively $60000$ and $10000$ labeled $28$-by-$28$ grayscale digits. For feature extraction, our net, followed by a ten-way softmax layer, was 
\begin{align*}
C(64, 3) \rightarrow C(64, 3) \rightarrow P(2) \rightarrow C(64, 3) \rightarrow P(2) \rightarrow FC(128) 
\end{align*}
where $C(64,3)$, $P(2)$ and $FC(128)$ denotes a layer of $64$ $3$-by-$3$ convolutional filters, a max-pooling layer of $2$-by-$2$ windows with stride $2$, and a fully connected layer of $128$-dimensional fan-out.

We ran experiments \textit{without data augmentation}. The digits were linearly scaled to make their intensity distribution of zero mean and unit variance. First, the net was trained for $500$ epochs with $\lambda=500$ and $\rho=0.1$ on the training data. Its resultant clustering accuracy was $0.9838$. Its test accuracy was even better (Tab. \ref{utable}). We then ran a fresh training on the full dataset. With $\lambda=100$ and $\rho=0.1$, the same training also yielded a good result (Tab. \ref{utable}). Note that IMSAT \cite{hu2017learning} used data augmentation.  

\begin{table}
	\caption{Accuracy of unsupervised discriminative learning on MNIST and Reuters}
  \centering
\ra{0.1}
  \begin{tabular}{lllllllll}
    \toprule
	  & \multicolumn{4}{c}{clustering} 
	  & \multicolumn{2}{c}{generalization (training)}\\
	  \cmidrule(r{8pt}){2-5} \cmidrule(lr){6-7}
	  &  VaDE \cite{jiang2016variational} & IMSAT \cite{hu2017learning} & S-Net \cite{shaham2018spectralnet} & Ours & S-Net \cite{shaham2018spectralnet} & Ours\\
	  \midrule
	  MNIST & 0.9446 & \textbf{0.988} & 0.972 & 0.9692 & 0.970 (n.a.) & \textbf{0.9878 (0.9838)}\\
	  \midrule
	  Reuters & 0.7938 & 0.719 & 0.809 & \textbf{0.8323} & 0.798 (n.a.) & \textbf{0.8094 (0.8107)} \\
    \bottomrule
  \end{tabular} \label{utable}
\end{table}

%

\begin{figure}
  \centering
	\includegraphics[width = 0.48\textwidth]{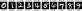} \hspace{0.3cm}
   \includegraphics[width = 0.48\textwidth]{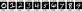}\\
	\vspace{0.1cm}
	\includegraphics[width = 0.48\textwidth]{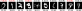} \hspace{0.3cm}
   \includegraphics[width = 0.48\textwidth]{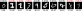}\\
	\vspace{0.1cm}
	\includegraphics[width = 0.48\textwidth, height = 0.12\textheight]{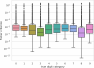}\hspace{0.4cm}
	\includegraphics[width = 0.48\textwidth, height = 0.12\textheight]{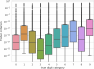}
	\caption{Comparison of a supervised (left column; test error $0.54\%$) and unsupervised (right column; test error $1.22\%$) model trained on the same set of $60000$ digits. The top two rows are \textit{test} digits of the smallest and largest Fisher criterion in their respective category under the two models. The unsupervised model's labeling was remapped by the Munkres algorithm. The caption $a (b)$ on top of a digit reads a model's prediction $a$ and its confidence $b$. Red caption indicates a mis-classified digit. The numbers beneath are their model specific Fisher criterion. A larger criterion indicates a stronger vulnerability to attacks. The bottom row is a box plot of the category-wise Fisher criterion. Our functional regularization lowers the average Fisher criterion for all categories.} \label{fig1}
\end{figure}

\begin{figure}
  \centering
   \includegraphics[height=0.1\textheight, width = 0.32\textwidth]{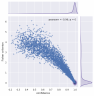}
   \includegraphics[height=0.1\textheight, width = 0.32\textwidth]{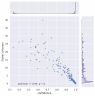}
   \includegraphics[height=0.1\textheight, width = 0.32\textwidth]{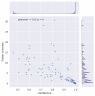}
\caption{Joint distributions of predictive confidence and Fisher criterion of the same net trained under different settings. We collected the statistics on the MNIST test set. From left to right are an overfitted model (trained on 100 labeled digits only), a supervised model trained on $60000$ labeled digits and an unsupervised model trained on the same digits but with no labels. The overfitted model is smooth but not confident whereas both well trained models are confident. The unsupervised model is even smoother.} \label{fig2}
\end{figure}

To better understand the effects of regularization in terms of model confidence and stability, we also trained the same net in two supervised settings, one on the full training set (test error $0.54\%$) and the other on only $100$ random labeled digits, $10$ per category (test error $11.07\%$). We recorded their predictive confidence and Fisher criterion at all the test digits. Fig. \ref{fig1} shows the most and least stable digits according to the two well trained models whereas Fig. \ref{fig2} illustrates the effect of regularization, either by additional data or by our functional constraints. It shows that our functional regularization lowers the average Fisher criterion for all categories and also attains high model predictive confidence. 



\subsection{Reuters}

Reuters \cite{lewis2004rcv1} is a labeled corpus of English news. Following the established practice \cite{xie2016unsupervised}\cite{jiang2016variational}\cite{hu2017learning}\cite{shaham2018spectralnet}, we took four categories i.e. corporate/industrial, government/social, markets and economics and computed normalized tf-idf features on the 2000 most frequent words. This preprocessing represents each document as a vector of squared 2-norm equal to $2000$. The resulting dataset was skewed with the least frequent category representing $7.98\%$ of its $685,071$ documents. It was randomly divided to a $90\%$-$10\%$ split with the larger subset used for training. We took the same net as in \cite{shaham2018spectralnet}, which has two fully connected hidden layers containing respectively $512$ and $256$ neurons for feature extraction. With $\lambda=100$ and $\rho = 0.02$, $150$ training epochs resulted in a net with clustering accuracy $0.8107$ and test accuracy $0.8094$. When trained on the full dataset with the same parameters, its clustering accuracy was even higher (Tab. \ref{utable}).

\section{Discussion} \label{sec:relation}
 
To conclude, in this paper, we have presented an unsupervised framework to constrain a probabilistic discriminative learner's hypothesis space to a set of non-trivial piecewise constant functions. This functional constraint enforces a learned model's predictive confidence and smoothness, allowing it to generalize. Our approach is generic in that it applies to all probabilistic discriminative learners and can be used for scalable unsupervised discriminative learning. Due to page limit, we now discuss several prior works which directly inspire our work and future research directions. 

\paragraph{Confidence regularization and spectral clustering.} First we state a result which underpins this part of the discussion. Consider a discriminative model $\mathbb{Q}(\cdot|x), x \in \mathcal{X}$ and a finite set of instances $\mathcal{S} \subseteq \mathcal{X}$, over which we define a Markov chain with the following \textit{instance transition matrix}
\begin{align}
\forall (x', x) \in \mathcal{S} \times \mathcal{S}, \quad \mathbb{S}(x'|x; \mathbb{Q}, \mathcal{S}) := \sum_{y \in \mathcal{Y}} \mathbb{P}(x'|y; \mathbb{Q}, \mathcal{S}) \mathbb{Q}(y|x). \label{mchain} 
\end{align}
Note that by the very definition of $\mathbb{P}(\cdot|y; \mathbb{Q}, \mathcal{S})$, i.e. Eq. \eqref{l2i}, this transition matrix is symmetric. The next theorem states the implication of a diagonal label transition matrix in terms of this Markov chain. 
\begin{thm}
The Markov chain has $|\mathcal{Y}|$ irreducible recurrent classes if and only if the label transition matrix $\mathbb{T}(\cdot|\cdot; \mathbb{Q}, \mathcal{S})$, i.e. Eq. \ref{labeltm}, is diagonal.\label{thm3}
\end{thm}
 
Spectral clustering is one of the most popular approaches to unsupervised clustering \cite{shi2000normalized} \cite{ng2002spectral} \cite{von2007tutorial}. For its application, an adjacency metric is needed to measure pairwise instance similarity. Here we use the transition matrix $\mathbb{S}(\cdot|\cdot; \mathbb{Q}, \mathcal{S})$ for its modeling and see under which conditions it can be made useful for clustering $|\mathcal{Y}|$ classes of instances. The answer is clearly when the matrix $\mathbb{S}$ leads to $|\mathcal{Y}|$ irreducible recurrent classes, which according to Theorem \ref{thm3} is equivalent to a diagonal label transition matrix. Moreover, as our confidence regularization allows the unlabeled set $\mathcal{S}$ to vary as long as it is label complete, it can be seen as a scalable approach to adjacency and spectral learning when training data is not too imbalanced (ref. the proof of Theorem \ref{thm3} in the Appendix). 

A recent work \cite{haeusser2017learning} explores a similar idea but it directly models the instance transition matrix. Unlike our approach, theirs does not explicitly specify the number of irreducible classes in data, hence the risk of over-fragmentation, i.e. leaving some data unvisited. The authors thus introduced an entropy loss to favor a uniform visit. Our model, by operating directly at the category level, avoids this issue. 

\begin{table}
\caption{Ideal loss under four discriminative training schemes}
  \centering
\ra{0.1}
  \begin{tabular}{llll}
    \toprule
	  \multicolumn{2}{c}{}\\
ord. supervised  &$L_o(\theta):=\mathbb{E}_{X\sim \mathbb{Q}^*}\kl(\mathbb{Q^*}(\cdot|X) || \mathbb{Q}(\cdot|X; \theta))$\\
	  adversarial \cite{goodfellow2014explaining} &$L_a(\theta;\rho):=\mathbb{E}_{X\sim \mathbb{Q}^*}\sup_{\|r\|_2 \leq \rho}\kl(\mathbb{Q^*}(\cdot|X) || \mathbb{Q}(\cdot|X+r; \theta))$\\
	  VAT  \cite{miyato2017virtual} &$L_v(\theta;\hat{\theta}, \rho):=\mathbb{E}_{X\sim \mathbb{Q}^*}\sup_{\|r\|_2 \leq \rho}\kl(\mathbb{Q}(\cdot|X; \hat{\theta}) || \mathbb{Q}(\cdot|X+r; \theta))$\\
	  smoothness reg. &$L'_s(\theta;\rho):=\mathbb{E}_{X\sim \mathbb{Q}^*}\sup_{\|r\|_2 \leq \rho}\hel^2(\mathbb{Q}(\cdot|X; \theta) || \mathbb{Q}(\cdot|X+r; \theta))$\\
    \bottomrule
  \end{tabular} \label{table2}
\end{table}
\paragraph{Smoothness regularization and virtual adversarial training.} They differ in formulation. Specifically, the former favors \textit{all} $\mathbb{Q}^*$-piecewise constant models in a hypothesis space \textit{equally} whereas the latter, as stated in \cite{miyato2017virtual}, is intended as an unsupervised alternative to adversarial training \cite{goodfellow2014explaining} with the goal of approximating $x\mapsto \mathbb{Q}^*(\cdot|x)$ itself only. Tab. \ref{table2} shows the ideal losses under various schemes.

Both unsupervised, the two are related. VAT, when searching for an optimal perturbation direction at instance $x$, implicitly solves for the leading eigenvector of the Fisher information $I_F(x, \hat{\theta})$. However, since this vector must lie in the subspace spanned by the score vectors, in high dimension, it is much more efficient to exploit its low-rankedness. Moreover, our study interprets VAT's power iteration as a sampling procedure. Given a Gaussian \textit{direction} $e$, we suggest it be used as a \textit{dimension} for a grid search, because it necessarily results in a tighter lower bound to $\phi_f(\cdot; x,\theta)$. Our formulation also gives $\rho$ a probabilistic meaning as an estimate of the $\mathbb{Q}^*$-dependent attack-free margin i.e. Eq. \eqref{radius}. 

In the future, we plan to study how to take noise into account in confidence regularization. For smoothness regularization, we will see whether an explicit instance manifold modeling would lead to a better result. Moreover, we consider applying our framework to other learning modes, such as semi-supervised learning \cite{haeusser2017learning}, active learning \cite{dasgupta2009two} and outlier detection.

\appendix
\section{Appendix} \label{sec:app} 

\textit{Proof (of Theorem \ref{thm1})} The {\em{if}} part. Consider a binary valued model $\mathbb{Q}(\cdot|x)$. For all $y \in \mathcal{Y}$, define the non-empty set $\mathcal{S}^{\mathbb{Q}}_y :=\{x \in \mathcal{S}\;|\; \mathbb{Q}(y|x) = 1\}$. It follows from Eq. \eqref{l2i} $\mathbb{P}(x|y; \mathbb{Q}, \mathcal{S}) = 1_{\mathcal{S}^{\mathbb{Q}}_y }(x) |\mathcal{S}_y^{\mathbb{Q}}|^{-1}$. Therefore, denoting by $\delta_{yy'}$ the Kronecker delta, we find 
\begin{align*}
\mathbb{T}(y'|y; \mathbb{Q}, \mathcal{S}) = \sum_{x \in \mathcal{S}} \mathbb{P}(x|y; \mathbb{Q}, \mathcal{S}) \mathbb{Q}(y'|x)
	= |\mathcal{S}_y^{\mathbb{Q}}|^{-1}\sum_{x \in \mathcal{S}_y^{\mathbb{Q}}} \mathbb{Q}(y'|x) = \delta_{yy'}.
\end{align*}
To show the \textit{only if} part, consider a diagonal label transition matrix $\mathbb{T}(\cdot|\cdot;\mathbb{Q}, \mathcal{S})$. Assume for some $y \in \mathcal{Y}$ such that $\{x \in \mathcal{S}, \;\mathbb{Q}(y|x) > 0\} = \emptyset$. Since $\mathbb{Q}(y|x)$ is valued in $[0,1]$, the equality
\begin{align*}
	\mathbb{T}(y|y; \mathbb{Q}, \mathcal{S}) = \sum_{x \in \mathcal{S}} \mathbb{P}(x|y; \mathbb{Q}, \mathcal{S}) \mathbb{Q}(y|x) = 1 
\end{align*}
implies that $\mathbb{Q}(y|x) = 1$ over $\mathbb{P}(\cdot|y; \mathbb{Q}, \mathcal{S})$'s support, which has to be non-empty, hence a contradiction. It follows $\{x \in \mathcal{S}, \;\mathbb{Q}(y|x) > 0\} \neq \emptyset$ for all $y\in \mathcal{Y}$. The same reasoning then implies
\begin{align*}
	\forall y \in \mathcal{Y},\quad \{x \in \mathcal{S}, \;\mathbb{Q}(y|x) > 0\} = \{x \in \mathcal{S}, \;\mathbb{Q}(y|x) = 1\}. \cqfd
\end{align*}

\textit{Proof (of Corollary \ref{col1})} In view of Theorem \ref{thm1}, it suffices to prove $\forall (x, y) \in \mathcal{X}\times\mathcal{Y}, \; \mathbb{Q}^*(y|x)\in\{0,1\}$. It is exactly what the noiseless condition implies. $\blacksquare$

\textit{Proof (of Theorem \ref{thm2})} Under the conditions stipulated in the theorem, the union bound leads to 
\begin{align*}
	\mathbb{Q}^*\left( \cup_{j=1}^T \cup_{y\in \mathcal{Y}} \{\mathcal{S}_j \cap \mathcal{X}_y = \emptyset\}\right)
	\leq T |\mathcal{Y}|\max_{y\in \mathcal{Y}, |\mathcal{S}|=b}\mathbb{Q}^*\left(\{\mathcal{S} \cap \mathcal{X}_y = \emptyset\}\right)
	\leq T |\mathcal{Y}|(1-\min_{y\in \mathcal{Y}}\mathbb{Q}^*(Y=y))^{b}.
\end{align*}
For any $\epsilon\in (0,1)$, the condition $b > -\ln \left(T |\mathcal{Y}|\epsilon^{-1}\right)/\ln\left(1-\min_{y\in \mathcal{Y}}\mathbb{Q}^*(Y=y)\right)$ thus implies
\begin{align*}
	\mathbb{Q}^*\left( \cap_{j=1}^T \cap_{y\in \mathcal{Y}} \{\mathcal{S}_j \cap \mathcal{X}_y \neq \emptyset\}\right)
	=1-\mathbb{Q}^*\left( \cup_{j=1}^T \cup_{y\in \mathcal{Y}} \{\mathcal{S}_j \cap \mathcal{X}_y = \emptyset\}\right) > 1-\epsilon.
\end{align*}
Finally, observe that $-\ln(1-x) \geq x$ holds whenever $x < 1$. The proof ends. $\blacksquare$



\textit{Proof (of Theorem \ref{thm3})} The \textit{if} part. Consider a diagonal $\mathbb{T}(\cdot|\cdot; \mathbb{Q}, \mathcal{S})$ and an arbitrary $y \in \mathcal{Y}$. We find 
\begin{align*}
	\sum_{x\in \mathcal{S}}\mathbb{S}(x'|x; \mathbb{Q}, \mathcal{S}) \mathbb{Q}(y|x) 
= \sum_{x\in \mathcal{S}} \sum_{y' \in \mathcal{Y}} \mathbb{P}(x|y'; \mathbb{Q}, \mathcal{S}) \mathbb{Q}(y'|x') \mathbb{Q}(y|x)
= \sum_{y' \in \mathcal{Y}} \mathbb{T}(y|y'; \mathbb{Q}, \mathcal{S}) \mathbb{Q}(y'|x').
\end{align*}
The first equality holds because $\mathbb{S}$ is symmetric. Therefore, for all $y \in \mathcal{Y}$, the measure $\mu_y(\cdot) := 1_{\mathcal{S}}(\cdot)\mathbb{Q}(y|\cdot)$ is stationary. Moreover, its support has to be part of an irreducible recurrent class because for any of its two elements, there is a path of non-zero probability to connect them. In addition, by Theorem \ref{thm1}, $\mu_y$ and $\mu_{y'}$ have disjoint supports whenever $y\neq y'$. Hence we find $|\mathcal{Y}|$ irreducible recurrent classes.

To prove the \textit{only if} part, note that as defined by Eq. \eqref{mchain}, the transition matrix has its rank upper bounded by $|\mathcal{Y}|$. As a result, its associated Markov chain has at most $|\mathcal{Y}|$ disjoint irreducible recurrent classes. For the Markov chain to have exactly $|\mathcal{Y}|$ disjoint recurrent classes, these supports thus must not overlap. Hence, they form a partition of $\mathcal{S}$ and the proof ends. $\blacksquare$

\bibliographystyle{plainnat}
\bibliography{mybib}
\end{document}